\pdfoutput=1

\documentclass[11pt]{article}

\usepackage[]{acl}
\usepackage{setspace}
\usepackage{times}
\usepackage{latexsym}
\usepackage{color}
\usepackage{tcolorbox}
\usepackage{colortbl}
\usepackage{CJK}
\usepackage{adjustbox}
\usepackage{multirow}
\usepackage{verbatim}
\usepackage[T1]{fontenc}

\usepackage[utf8]{inputenc}

\usepackage{microtype}

\usepackage{tikz}
\def\checkmark{\tikz\fill[scale=0.25](0,.35) -- (.25,0) -- (1,.7) -- (.25,.15) -- cycle;}

\usepackage{graphicx}
\newcommand{\minus}{\scalebox{0.75}[1.0]{$-$}}

\usepackage{algorithm,algorithmic}

\newcommand{\prt}[1]{\text{p}_\theta\left( #1\right)}

\newcommand{\eg}{\textit{e.g.}}

\usepackage{amssymb}
\usepackage{amsmath}
\usepackage{}
\usepackage{esvect}
\usepackage{enumitem}
\usepackage{setspace}

\usepackage{dblfloatfix}

\title{Why Is It Hate Speech? \\Masked Rationale Prediction for Explainable Hate Speech Detection}

\author{Jiyun Kim, Byounghan Lee, Kyung-Ah Sohn\(^{*}\) \\
  Ajou University \\
  \texttt{\{hamjee66, qudgks96, kasohn\}@ajou.ac.kr} \\
  \(^{*}\)Corresponding author \\
  }
\begin{document}
\maketitle


\begin{abstract}

In a hate speech detection model, we should consider two critical aspects in addition to detection performance--bias and explainability. Hate speech cannot be identified based solely on the presence of specific words: the model should be able to reason like humans and be explainable. To improve the performance concerning the two aspects, we propose Masked Rationale Prediction (MRP) as an intermediate task. MRP is a task to predict the masked human rationales--snippets of a sentence that are grounds for human judgment--by referring to surrounding tokens combined with their unmasked rationales. As the model learns its reasoning ability based on rationales by MRP, it performs hate speech detection robustly in terms of bias and explainability. The proposed method generally achieves state-of-the-art performance in various metrics, demonstrating its effectiveness for hate speech detection. \textit{Warning: This paper contains samples that may be upsetting.}

\end{abstract}

\section{Introduction}

With the recent development of social media and online communities, hate speech, one of the critical social problems, can spread easily. The spread of hate strengthens discrimination and prejudice against the target social groups and can violate their human rights. Moreover, online hatred extends offline and causes real-world crimes. Therefore, properly regulating online hate speech is important to address many social problems related to aversion.  

In addition to the detection performance, two essential considerations are involved in implementing a hate speech detection model--\textit{bias} and \textit{explainability}. Hate speech should not be judged by any specific word but by the context in which the word is used. Even if any word generally considered vicious does not exist in a text, the text can be hate speech. A specific expression does not always imply hatred either (\eg, e.g., `nigger') \citep{del2017hate}. However, the presence of this word can cause a model to make a biased detection of hate speech. This erroneous judgment may inadvertently strengthen the discrimination against the target group of the expression \citep{sap2019risk, davidson2019racial}. In this respect, the model's \textit{bias} toward specific expressions should be excluded.

\begin{figure}[!t]
    \centering
    \includegraphics[width=1.0\linewidth]{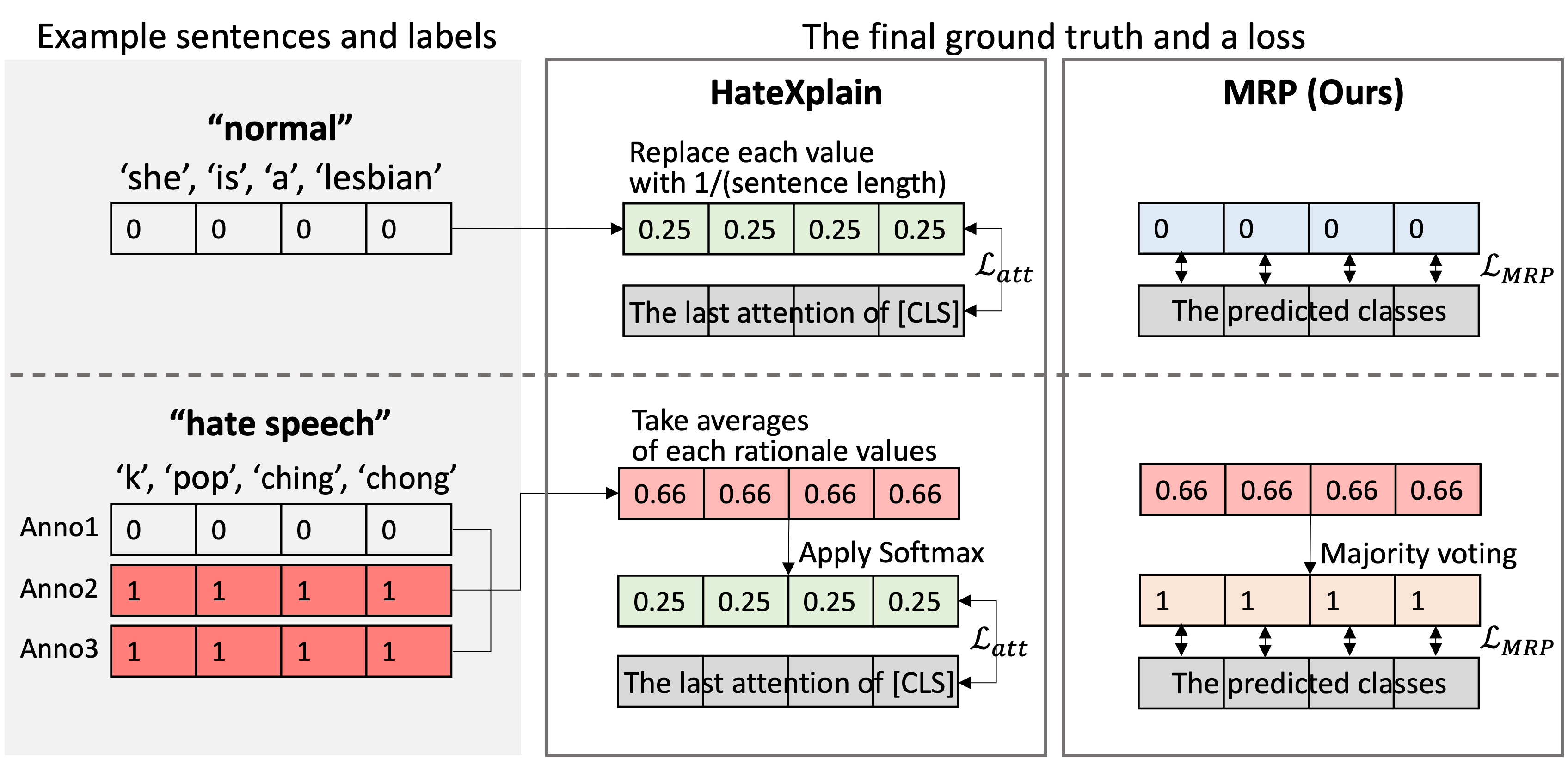}
    \caption{Examples for the two methods to get the final ground truths. Example input sentences are represented with the class and human rationale labels. In this figure, HateXplain uses the same ground truth about both normal and hateful sentences for the loss. However, our method could determine the two classes with the ground truths.}
    \label{fig:my_label}
    \label{fig:long}
    \label{fig:onecol}
\label{hatexplain_mech}
\end{figure}

\begin{figure*}
\centering 
\includegraphics[width=1.0\linewidth]{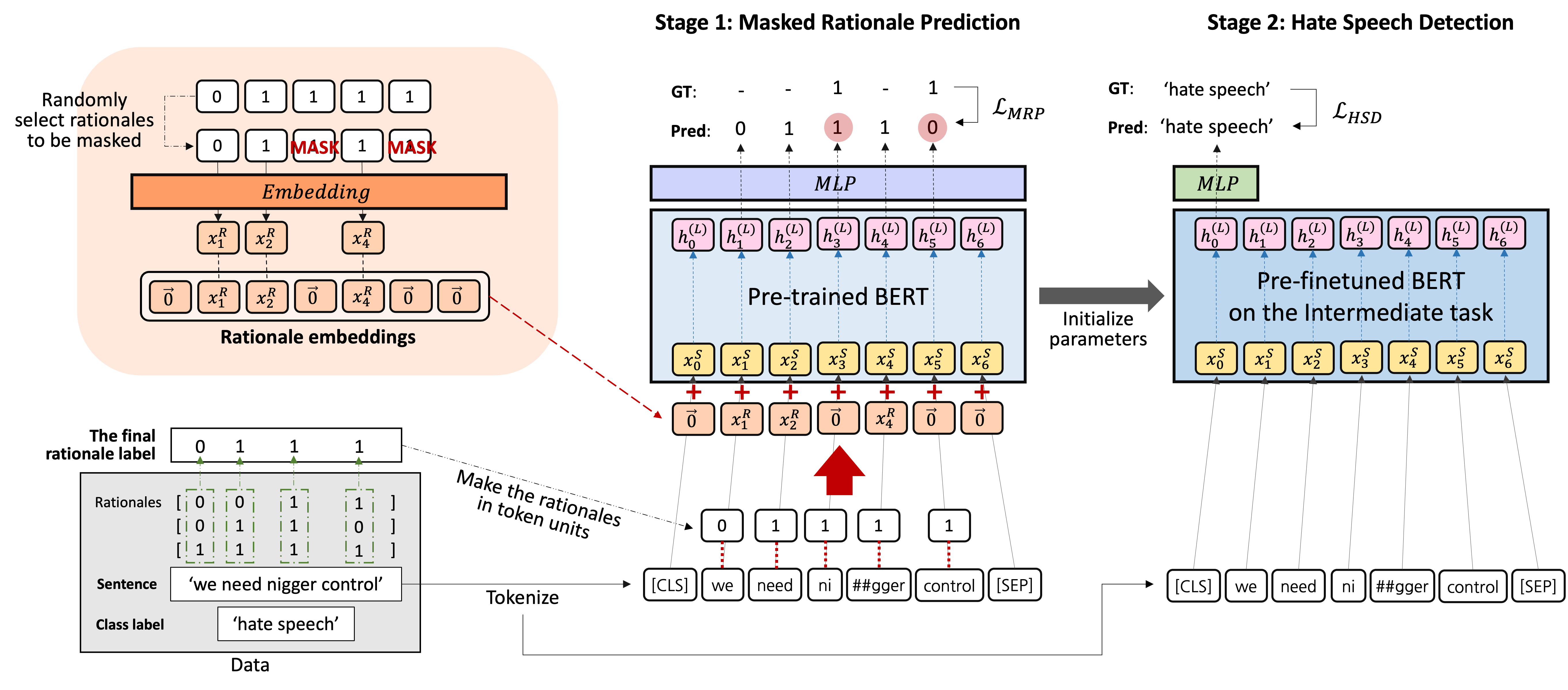}
\caption{Framework of the proposed method. We finetune a pre-trained BERT through two training stages--Masked Rationale Prediction (MRP) and then hate speech detection. In MRP, the partially masked rationale label is inputted as the rationale embeddings by being added into the input embeddings of BERT. The model predicts each masked rationale per token. The model for hate speech detection is initialized by the updated parameters during MRP. }
\label{model_architecture}
\end{figure*}

The expressions that can cause biased judgment should be interpreted in context. It means it is vital for the hate speech detection models to have the ability to make judgments based on context, as humans do. Therefore, the model should be \textit{explainable} to humans 
so that the rationale behind a result is explained \citep{liu2018towards}. Here, the rationale is a piece of a sentence as justification for the model's prediction about the sentence, as defined by related research \citep{hancock2018training, lei2016rationalizing}. 

To the best of our knowledge, HateXplain \citep{mathew2020hatexplain} is the first hate speech detection benchmark dataset that considers both these aspects. They proposed a method that utilizes rationales as attention ground truths to complement the performance of the two elements. 
However, when most tokens are annotated as the human rationale in a hateful sentence, the rationale's information could be meaningless as the ground truth attention becomes hard to be distinguished from that of a normal sentence, as shown in Figure~\ref{hatexplain_mech}. This can hinder the model's learning. 


In this paper, we present a method to implement a hate speech detection model much more effectively by using the human rationale of hate for finetuning a pre-trained language model. To achieve this, we propose Masked Rationale Prediction (MRP) as an intermediate task before finetuning the model on hate speech detection. MRP trains a model to predict the human rationale label of each token by referring to the context of the input sentence. The model takes the human rationale information of some input tokens among the sentence along with the corresponding tokens as input. It then predicts the rationale of the remaining tokens on which the rationale is masked. We embed the rationales to provide the human rationales as input per token. The masking process of the partial rationales is implemented while creating rationale embeddings; some rationale embedding vectors are replaced with zero vectors. 

MRP allows the model to make judgments per token about its masked rationale by considering surrounding tokens with an unmasked rationale. With this, the model learns a human-like reasoning process to get context-dependent abusiveness of tokens. The model parameters trained on MRP become the initial parameter values for hate speech detection in the following training stage. In this way, based on the way of human reasoning for hate, the model can get improved abilities in terms of bias and explainability in detecting hate speech. We experimented with BERT \cite{devlin2018bert} as the pre-trained model. Consequently, our models finetuned in the proposed way--BERT-MRP and BERT-RP--achieve state-of-the-art performance overall on all three types of 11 metrics of HateXplain benchmark--Performance-based, Bias-based, and Explainability-based \citep{mathew2020hatexplain}. And the two models, especially BERT-MRP, also show the best results in qualitative evaluation of explicit and implicit hate speech detection. 

The main contributions of this paper are:
\begin{itemize}
    \item We propose a method to utilize human rationales as input by transforming them into rationale embeddings. Combining the embedded rationales with the corresponding input sentence can provide information about the human rationales per token during model training.
    \item We propose Masked Rationale Prediction (MRP), a learning method that leads the model to predict the masked rationale by considering the surrounding tokens. The model is allowed to learn the reasoning process in context.
    \item We finetune a pre-trained BERT in two stages--on MRP as an intermediate task and then on hate speech detection. The parameters trained concerning human reasoning for hate become a sufficient basis not only for the detection but also in terms of the model bias and explainability.
\end{itemize}

\section{Related works}
\noindent
\textbf{Hate speech detection}   
With the advance of deep learning, hate speech detection studies have utilized neural networks \citep{badjatiya2017deep, han2019unsupervised}, and word embedding methods \citep{mckeown2018predictive}. More recently, Transformer-based \citep{vaswani2017attention} models have shown remarkable results. In hate speech detection, BERT has been adopted for various studies as hate speech detection can be considered a classification task. \citet{mandl2019overview} and \citet{ranasinghe2019brums} compared a BERT-based model with Recurrent Neural Networks (RNNs)-based models and showed the BERT-based model outperforms other models. Furthermore, some studies have considered the model's bias and explainability. \citet{vaidya2020empirical} improved accuracy and reduced unintended bias by adopting multi-task learning that predicts toxicity of text and target group labels as additional information. \citet{mathew2020hatexplain} utilized rationales of the dataset as additional information for finetuning BERT to deal with the bias and explainability. To improve performance in terms of the two considerations, we propose a more effective finetuning approach based on BERT and the rationales by adopting the pre-training framework.

\noindent
\textbf{Pre-finetuning on an intermediate task} 
Recently, finetuning a pre-trained model on a downstream task has become the norm~\citep{howard2018universal, radford2018improving}. However, it cannot be guaranteed that the model finetuned with a small dataset compared to its size will be sufficiently well-adjusted for the target downstream task~\citep{phang2018sentence}. Pre-finetuning is a technique to train the model on a task before the target task \citep{aghajanyan2021muppet}. This can help the model learn the data patterns or reduce the tuning time so that it converges quickly to better fit the target task. According to \citet{pruksachatkun2020intermediate, aghajanyan2021muppet}, the more closely the intermediate task is related to the target task, the better the effect of pre-finetuning. And inference tasks involving the reasoning process show a remarkable improvement in the target task performance. We adopt this method to train a pre-trained language model through two stages for hate speech detection. As the intermediate task, we propose MRP, which guides the model to infer the human rationale of each token based on surrounding tokens.  \\
\noindent
\textbf{Explainable NLP and rationale}  
Explaining the rationale of the result of an AI model is necessary for it to be explainable to humans \citep{liu2018towards}. Some Natural Language Processing (NLP) studies define rationale as snippets of an input text on which the model's prediction is supported \citep{hancock2018training, lei2016rationalizing}. \citet{lei2016rationalizing} implemented a generator that generates words considered rationales and used them as input of an encoder for sentiment classification. \citet{bao2018deriving} mapped the human rationales into the model attention values to solve the low-resource problem by learning a domain-invariant token representation. For hate speech detection, HateXplain employs the human rationales as ground truth attention to concentrate on aggressive tokens. Unlike existing approaches, we utilize the masked human rationale label embeddings as input. They become the useful additional information of each token. \\
\noindent
\textbf{Masked label prediction}    
The UniMP model presented by \citet{shi2020masked} aims to solve the graph node classification problem using graph transformer networks~\citep{yun2019graph}. 
They maximized the propagation information required to reconstruct a partially observable label by using both feature information and label information as inputs. However, to prevent overfitting due to excessive information, some label information is masked, and the masked label is predicted. We apply a similar method to text data for an intermediate task with rationales. Through additional rationale information, the model increases the understanding of input sentences, and the performance of the downstream task is improved.

\section{Method}
Hate speech detection can be described as a text classification problem. Following the problem setting of HateXplain, we define the problem as a three-class classification involving three categories--`hate speech,' `offensive,' or `normal'. We finetune a pre-trained BERT on hate speech detection. Note that other transformer encoder-based models can be used instead. Before finetuning the model on this task, we pre-finetune it on an intermediate task. We propose Masked Rationale Prediction (MRP) as the intermediate task. Our method is described in Figure~\ref{model_architecture}.

\subsection{Masked rationale prediction}
For MRP, we utilize human rationales of hate provided by the HateXplain dataset. Annotators marked some words in a sentence as rationales for judging the sentence as abusive. A rationale label is presented in a list format, including 1 as rationale and 0 as non-rationale per word in the corresponding sentence. There are no such labels for a sentence whose final class is `normal.' As the dataset was annotated by two or three people per sentence, some pre-processing is required to get the final rationale labels for MRP. To manipulate the multiple rationale labels to one per sentence, we obtain the average value of the rationales per word, and if it is over 0.5, the value of 1; otherwise, the value of 0 is determined as the final rationale of the corresponding word. The final rationale label is a list of these last values.
In the case of the `normal' sentence, a list of zeros is used. Accordingly, the final rationale label consists of as many 0s or 1s as the number of words in the sentence. As a sentence is tokenized, its rationale label is also modified in token units.

MRP is based on token classification, which predicts the rationale label $R$ per token in an input sentence $S$. In our MRP, the rationale labels, as well as the sentences, are used as inputs. The process of embedding  $S$ is the same as that of BERT. We denote the embedded $S$ as $X^S=\{x_0^S, x_1^S, \cdots, x_{n-1}^S\} \in \mathbb{R}^{n\times d}$ where $n$ is the sequence length and $d$ is the embedding size. And to use $R$ as input, we pass it through an embedding layer to get $X^R=\{x_0^R, x_1^R, \cdots, x_{n-1}^R\} \in \mathbb{R}^{n\times d}$ as shown in Figure~\ref{model_architecture}.
The rationale embeddings reflect the attributes of each token as a ground for the human judgment.

MRP differs from BERT's Masked Language Modeling (MLM) in masking processing. Specifically, we do not mask the tokens; we mask the rationales. To construct the partially masked rationale embeddings $\Tilde{X}^R$, some rationales are randomly selected to be masked. Each of rationales is transformed into its corresponding embedding vector, except the masked ones. For masking, zero vectors replace the embedding vectors of each corresponding token. For example, if we mask $x_1^R$ and $x_3^R$, then the rationale embedding matrix is like $\Tilde{X}^R=\{\vv{0}, \vv{0}, x_2^R, \vv{0}, \cdots, x_{n-2}^R, \vv{0}\}$. The first and last rationale embeddings corresponding to CLS and SEP tokens, respectively, are replaced with $\vv{0}$.

The MRP model predicts the rationale by taking the sum of the embedded tokens $X^S$ and the partially masked rationales $\Tilde{X}^R$ as input. We then get:
\begin{equation}
\begin{split}
    \displaystyle
    H_{MRP}^{(0)} &= X^S + \Tilde{X}^R, \\
    H_{MRP}^{(l+1)} &= \text{Transformer}(H_{MRP}^{(l)}), \\  
    \hat{X}^R &= \text{MLP}(H_{MRP}^{(L)}).
\end{split}
\label{eq_mrp_forward}
\end{equation}
The $l$-th hidden state passes through the transformer block to create the $l+1$-th hidden state, and the last hidden state $H_{MRP}^{(L)}$ outputs a predicted rationale $\hat{X}^R$ through Multi-Layer Perceptron (MLP). In other words, the model is guided to predict the masked rationales by referring to the representations of tokens using their corresponding observed rationales.

The loss $\mathcal{L}_{MRP}$ is calculated with only the predictions of the masked rationales. Therefore, our objective function is:
\begin{equation}
\begin{split}
    \displaystyle
    \arg \max_{\theta} \hspace{1mm} \log \prt{\hat{X}^R \vert X^S, \Tilde{X}^R} = \\ 
    \sum_{m \in M}^{}\log \prt{x_m^R \vert X^S, \Tilde{X}^R}, \\
\end{split}
\label{eq_mrp_obj}
\end{equation}
where $M$ indicates a set of index numbers of rationales that have been masked.

\begin{table*}[!htb]
\centering
{\small
\begin{tabular}{l|cc|ccc|ccc}
\hline
\multicolumn{3}{c}{\textbf{Model}} & \multicolumn{3}{c}{\textbf{Performance}} & \multicolumn{3}{c}{\textbf{Bias}} \\
& ration. & pre-fin. & Acc. & Macro F1 & AUROC & GMB-Sub. & GMB-BPSN & GMB-BNSP \\
BERT & & & 69.0 & 67.4 & 84.3 & 76.2 & 70.9 & 75.7 \\
BERT-HateXplain & \checkmark & & 69.8 & 68.7 & 85.1 & 80.7 & 74.5 & 76.3 \\
BERT-MLM & & \checkmark & 70.0 & 67.5 & \underline{85.4} & 79.0 & 67.7 & 80.9 \\
\hline
BERT-RP & \checkmark & \checkmark & \textbf{70.7} & \underline{69.3} & 85.3 & \underline{81.4} & \underline{74.6} & \underline{84.8} \\
BERT-MRP & \checkmark & \checkmark & \underline{70.4} & \textbf{69.9} & \textbf{86.2} & \textbf{81.5} & \textbf{74.8} & \textbf{85.4} \\
\hline
\end{tabular}
}
\caption{Results for the performance-based and the bias-based metrics. Scores in bold type are the best for each corresponding metric, while the underlined are the second best, and so are in Table~\ref{explainability}. }
\label{perform_n_bias}
\end{table*}

\begin{table*}[!htb]
\centering
{\small
\begin{tabular}{l|cc|ccc|cc}
\hline
\multicolumn{3}{c}{\textbf{Model}}  & \multicolumn{5}{c}{\textbf{Explainability}} \\
\multicolumn{3}{c}{} & \multicolumn{3}{c}{\textbf{Plausibility}} & \multicolumn{2}{c}{\textbf{Faithfulness}} \\
& ration. & pre-fin. & IOU F1 & Token F1 & AUPRC & Comp. & Suff. $\downarrow$ \\
BERT [Att] & & & 13.0 & 49.7 & \underline{77.8} & 44.7 & 5.7 \\
BERT [LIME] & & & 11.8 & 46.8 & 74.7 & 43.6 & 0.8 \\
BERT-HateXplain [Att] & \checkmark & & 12.0 & 41.1 & 62.6 & 42.4 & 16.0 \\
BERT-HateXplain [LIME] & \checkmark & & 11.2 & 45.2 & 72.2 & \textbf{50.0} & 0.4 \\
BERT-MLM [Att] & & \checkmark & 13.5 & 43.5 & 60.8 & 40.1 & 11.9 \\
BERT-MLM [LIME] & & \checkmark & 11.3 & 47.2 & 76.5 & 43.4 & \textbf{-5.5} \\
\hline
BERT-RP [Att] & \checkmark & \checkmark & \underline{13.8} & \underline{50.3} & 73.8 & 45.4 & 7.2 \\
BERT-RP [LIME] & \checkmark & \checkmark & 11.4 & 49.3 & 77.7 & \underline{48.6} & \underline{-2.6} \\
BERT-MRP [Att] & \checkmark & \checkmark & \textbf{14.1} & \textbf{50.4} & 74.5 & 47.9 & 6.7 \\
BERT-MRP [LIME] & \checkmark & \checkmark & 12.9 & 50.1 & \textbf{79.2} & 48.3 & -1.2 \\
\hline
\end{tabular}
}
\caption{Results for the explainability-based metrics. The lower the score Sufficiency in Faithfulness, the better, and the higher the other scores, the better. }
\label{explainability}
\end{table*}

\subsection{Hate speech detection}

Hate speech detection is implemented as  three-class text classification. The model predicts which category $Y$ the input sentence belongs to  among `hate speech', `offensive', and `normal'. The head that outputs the predicted class $\hat{Y}$ is used on the top of BERT. Before training, the model parameters are initialized by parameters updated on the intermediate task MRP, except for the head. As the forms of heads are different for two stages, their parameters  are randomly initialized. Consequently, in the finetuning stage on hate speech detection, the rationale labels are not involved functionally, considered as $[0]_{n\times d}$. Therefore, in this stage, the input of the model is $H_{HSD}^{(0)} = X^S$.

In this stage, the model does not refer to the rationale labels. The parameters trained during MRP are utilized as a base for reasoning hatefulness in context. The loss $\mathcal{L}_{HSD}$ is obtained through a cross-entropy function, as the task is a multi-class classification problem.
\begin{equation}
\begin{split}
    \displaystyle
    \arg \max_{\theta} \hspace{1mm} \log \prt{\hat{Y} \vert X^S}.
\end{split}
\label{eq_hsd_obj}
\end{equation}

\section{Experiments}
\subsection{Dataset}
For both stages of the intermediate and the target task, we use the HateXplain dataset. It contains 20,148 items collected from Twitter and Gab. Every item consists of one English sentence with its own ID and annotations about labels for its category, target groups, and rationales, which are annotated by two or three annotators. Based on the IDs, the dataset is split into 8:1:1 for training, validation, and test phases. Following the permanent split provided by the dataset, the models can't reference any test data during the training phases of all stages.

\subsection{Metrics}


The evaluation is according to the metrics of HateXplain, which are classified into three types: performance-based, bias-based, and explainability-based. The performance-based metrics measure the detection performance in distinguishing among three classes (i.e., hate speech, offensive, and normal). Accuracy, macro F1 score, and AUROC score are used as the metrics.

The bias-based metrics evaluate how biased the model is for specific expressions or profanities easily assumed to be hateful. HateXplain follows AUC-based metrics developed by \citet{borkan2019nuanced}. The model classifies the data into ‘toxic’--hateful and offensive--and ‘non-toxic’--normal. For evaluating the model’s prediction results, the data are separated into four subsets: $D_{g}^{+}, D_{g}^{-}$, $D^{+}$, and $D^{-}$. The target group labels are considered standard for dividing data into subgroups. The notations with $g$ denote the data of a specific subgroup among the subgroups, and the notations without g are the remaining data. $+$ and $-$ mean that the data are toxic and non-toxic, respectively. Based on these subsets, three AUC metrics are calculated.

Subgroup AUC is to evaluate how biased the model is to the context of each target group: $AUC(D_{g}^{-} + D_{g}^{+})$.
The higher the score, the less biased the model is with its prediction of a certain social group. 

BPSN (Background Positive, Subgroup Negative) AUC measures the model's false-positive rates regarding the target groups: $AUC(D^{+} + D_{g}^{-})$. 
The higher the score is, the less a model is likely to confuse non-toxic sentences whose target is the specific subgroup and toxic sentences whose target is one of the other groups. 

BNSP (Background Negative, Subgroup Positive) AUC measures the model's false-negative rates regarding the target groups: $AUC(D^{-} + D_{g}^{+})$. The higher the score is, the less the model is likely to confuse non-toxic sentences whose target is the specific group and toxic sentences whose target is one of the other groups. 

We calculate GMB (Generalized Mean of Bias)\footnote[1]{https://www.kaggle.com/competitions/jigsaw-unintended-bias-in-toxicity-classification/overview/evaluation} 
of the three metrics as the final scores to combine those ten scores of each of the metrics into one overall measure according to the HateXplain benchmark.
The formula is: $M_{p}(m_{s}) = (\frac{1}{N}\sum_{s=1}^{N}m_{s}^{p})^\frac{1}{p}$, where $M_{p}$ means the $p^{th}$ power-mean function, $m_{s}$ is one of the bias metrics $m$ calculated for a specific subgroup $s$, and N is the number of subgroups which is 10 in this paper.


The explainability-based metrics evaluate how much the model is explainable. HateXplain follows ERASER \citep{deyoung2019eraser}, which is a benchmark for the evaluation of explainability of an NLP model based on rationales. The metrics are divided into Plausibility and Faithfulness. Plausibility refers to how the model's rationale matches the human rationale. Plausibility can be considered both discrete selection and soft selection. For discrete selection, We convert token scores to binary values by more than some threshold(here 0.5). Then, We measures IOU F1 score and Token F1 score. For soft selection, We constructed AUPRC by sweeping a threshold over token scores.

Faithfulness evaluates the influence of the model rationale on its prediction result and consists of Comprehensiveness and Sufficiency. Comprehensiveness assumes the model prediction is less confidence when rationales are removed. This metric can be calculated: $m(x_{i})_{j} - m(x_{i} \backslash  r_{i})_{j}$. 
$m(x_{i})_{j}$ is the prediction probability of the corresponding class j with an input sentence $x_{i}$ by the model $m$. And $x_{i} \backslash  r_{i}$ is the sentence manipulated by removing the predicted rationale tokens $r_{i}$ from $x_{i}$.\footnote[2]{We select the top 5 tokens to remove based on the average length of human-annotated rationale labels in the dataset according to HateXplain benchmark.} The higher a score, the more influential the model's rationales in its prediction.
Sufficiency captures the extent to which extracted rationales are acceptable for a model to make a prediction: $m(x_{i})_{j} - m(r_{i})_{j}$. A low score of this metric means that the rationales are adequate in the prediction.

In addition, for the HateXplain benchmark, the scores are calculated based on the attention scores of the last layer or by using the LIME method \citep{ribeiro2016should}. The former is marked as [Att], and the latter is [LIME] in Table~\ref{explainability}. \citet{deyoung2019eraser} and \citet{mathew2020hatexplain} contain more detailed explanations. 

\subsection{Models and Experimental settings}

The evaluated models in Table~\ref{perform_n_bias} and Table~\ref{explainability} are as follows. All models are based on a BERT-base-uncased model for a pre-trained model and finetuned on hate speech detection. \textbf{BERT} in the tables is simply finetuned on hate speech detection with a fully-connected layer as a head for the three-class classification described above.

\textbf{BERT-HateXplain} uses attention supervision in addition to BERT. It matches the last attention values corresponding to the CLS token to the rationale used as ground truth attention. With this, the CLS token takes additional rationale-based attention information for the prediction. The loss is the summation of this attention loss and the detection loss. The results of BERT and BERT-HateXplain are the same as those presented in \citet{mathew2020hatexplain}.

\begin{figure}[!t] 
    \begin{center}
    \includegraphics[width=0.75\linewidth]{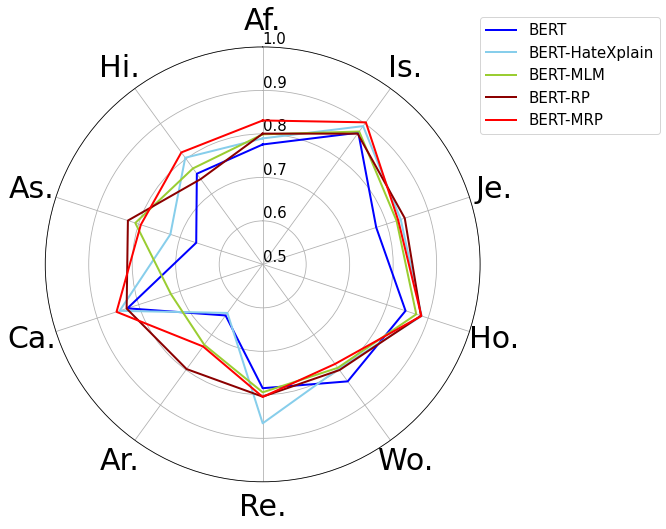}
    \end{center}
    \caption{The Subgroup scores among bias-based metrics for each of ten target groups. The target group labels are `African', `Islam', `Jewish', `Homosexual', `Women', `Refugee', `Arab', `Caucasian', `Asian', and `Hispanic' in clockwise direction respectively. The BPSN and BNSP scores are attached in Appendix.}
    \label{fig:long}
    \label{fig:onecol}
\label{bias_graph_sub}
\end{figure}

\begin{figure}[!t]
    \centering
    \includegraphics[width=1.0\linewidth]{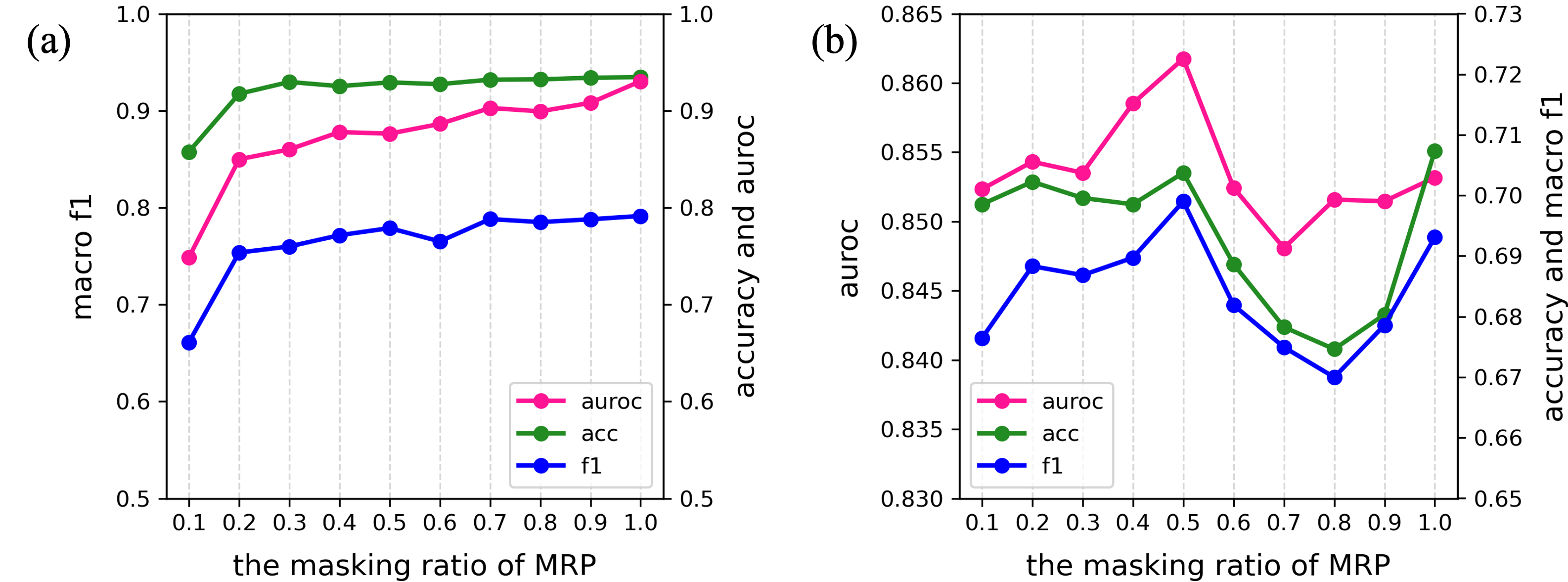}
    \caption{Classification test scores of the proposed models according to masking ratio in MRP. (a) is for token classification after training on MRP in the first stage, and (b) is for hate speech detection in the final stage. The case of masking 100\% of tokens is the same as BERT-RP.}
    \label{fig:my_label}
    \label{fig:long}
    \label{fig:onecol}
\label{ratio_graph}
\end{figure}

\textbf{BERT-MLM} is evaluated to compare the effectiveness of pre-finetuning tasks. Training a pre-trained NLP model with MLM using data of the downstream task is frequently used for the model to understand the downstream data and improve its performance \citep{han2019unsupervised, ben2020perl, arefyev2021nb}. It is implemented by simply masking 15\% tokens of each input sentence.

\textbf{BERT-MRP} and \textbf{BERT-RP} are the proposed models in this paper. \textbf{BERT-MRP} is the model trained on MRP as an intermediate task and then finetuned on hate speech detection. The ratio of masked rationales per token is set to 50\% of the entire rationale label. \textbf{BERT-RP} is trained on Rationale Prediction (RP), which is MRP when the ratio is set to 100\%--masking all the rationales. It is functionally the same as token classification with the rationale label as ground truth.

BERT-MLM, BERT-RP, and BERT-MRP are directly trained in this study. The experimental settings are the same for all models and each training step. The learning rate is $5e\minus 5$ during pre-finetuning and $2e\minus 5$ for hate speech detection, which is the same as BERT-HateXplain. We use the RAdam optimizer and an Nvidia GeForce GTX 1050 graphics card.

\subsection{Comparisons of results}

Table~\ref{perform_n_bias} and Table~\ref{explainability} present the performances of the models. For all metrics, the proposed models--the two from the bottom--perform much better overall.

\noindent
\textbf{Performance-based metrics}  As summarized in Table~\ref{perform_n_bias}, the proposed method outperforms the other methods. BERT-MRP shows the highest scores for Macro F1 and AUROC and BERT-RP for accuracy. The pre-finetuned models perform better than those that are not. It shows that the pre-finetuning process helps understand the data and allows enough time for tuning parameters for the target task, thereby improving performance. On the other hand, among the pre-finetuned models, the proposed models show better results than BERT-MLM. Furthermore, they outperform BERT-HateXplain, which also uses the rationale during training like ours. This shows that predicting the human rationale for hate as an intermediate task effectively implements a hate speech detection model.

\begin{table*}[ht!]
\setstretch{1.2}
\renewcommand{\arraystretch}{1.3}
\renewcommand{\tabcolsep}{0.7mm}
\centering
{\small
\begin{tabular}{lllr}
\hline
\textbf{No.} & \textbf{Model} & \textbf{Example sentence and Rationale} & \textbf{Label} \\
\hline
\hline

\hline
\multicolumn{4}{l}{\textbf{Case 1  Explicit hate speech}} \\
\hline
1 & Human & {\begingroup\setlength{\fboxsep}{0pt} \colorbox{red!0.0}{\strut imagine} \colorbox{red!0.0}{\strut unironically} \colorbox{red!0.0}{\strut believing} \colorbox{red!0.0}{\strut the} \colorbox{red!0.0}{\strut high} \colorbox{red!0.0}{\strut iq} \colorbox{red!100.0}{\strut kike} \colorbox{red!0.0}{\strut meme} \endgroup} & HS \\
\hline
 
 & BERT-MRP & {\begingroup\setlength{\fboxsep}{0pt} \colorbox{red!1.0006131156096827}{\strut imagine} \colorbox{red!0.5065305742354284}{\strut un} \colorbox{red!0.0}{\strut \#\#iro} \colorbox{red!0.4237702027203226}{\strut \#\#nical} \colorbox{red!1.1525533990086099}{\strut \#\#ly} \colorbox{red!1.5751172363058914}{\strut believing} \colorbox{red!4.611834668847015}{\strut the} \colorbox{red!2.7007141431724007}{\strut high} \colorbox{red!22.994687762516953}{\strut iq} \colorbox{red!100.0}{\strut ki} \colorbox{red!45.99515834398427}{\strut \#\#ke} \colorbox{red!8.685800375923666}{\strut me} \colorbox{red!3.8690364100821335}{\strut \#\#me} \endgroup} & HS \\
 
 & BERT-RP &  {\begingroup\setlength{\fboxsep}{0pt} \colorbox{red!0.7034893184091628}{\strut imagine} \colorbox{red!0.7963845283595737}{\strut un} \colorbox{red!0.8075214924718617}{\strut \#\#iro} \colorbox{red!0.47719949438441206}{\strut \#\#nical} \colorbox{red!3.067917779261801}{\strut \#\#ly} \colorbox{red!2.126502693141229}{\strut believing} \colorbox{red!0.9143983076525116}{\strut the} \colorbox{red!0.5038244526588006}{\strut high} \colorbox{red!0.4995158416455613}{\strut iq} \colorbox{red!100.0}{\strut ki} \colorbox{red!53.90257614482291}{\strut \#\#ke} \colorbox{red!0.2285388881856513}{\strut me} \colorbox{red!0.0}{\strut \#\#me}\endgroup} & HS \\
 
 & BERT-HX & {\begingroup\setlength{\fboxsep}{0pt} \colorbox{red!6.750447695330773}{\strut imagine} \colorbox{red!0.0}{\strut un} \colorbox{red!1.4681769078212856}{\strut \#\#iro} \colorbox{red!1.845767459618322}{\strut \#\#nical} \colorbox{red!2.9785060720637473}{\strut \#\#ly} \colorbox{red!7.220041451620094}{\strut believing} \colorbox{red!7.535339756906101}{\strut the} \colorbox{red!3.5315637915961338}{\strut high} \colorbox{red!14.10186428000596}{\strut iq} \colorbox{red!100.0}{\strut ki} \colorbox{red!86.57163891023302}{\strut \#\#ke} \colorbox{red!9.6692557068145}{\strut me} \colorbox{red!28.224579130781365}{\strut \#\#me} \endgroup} & HS \\
\hline

2 & Human & {\begingroup\setlength{\fboxsep}{1pt} \colorbox{red!0.0}{\strut <user>} \colorbox{red!0.0}{\strut ffs} \colorbox{red!0.0}{\strut these} \colorbox{red!100.0}{\strut ghetto} \colorbox{red!100.0}{\strut blacks} \colorbox{red!0.0}{\strut are} \colorbox{red!50.0}{\strut ignorant} \colorbox{red!0.0}{\strut like} \colorbox{red!100.0}{\strut cracker} \colorbox{red!100.0}{\strut mountain} \colorbox{red!100.0}{\strut hillbillies}  \endgroup}  & HS \\
\hline

 & BERT-MRP &  {\begingroup\setlength{\fboxsep}{1pt} \colorbox{red!0.0}{\strut <user>} \colorbox{red!0.0}{\strut ff} \colorbox{red!0.0}{\strut \#\#s} \colorbox{red!0.0}{\strut these} \colorbox{red!43.647928082874785}{\strut ghetto} \colorbox{red!100.0}{\strut blacks} \colorbox{red!42.1722411635775}{\strut are} \colorbox{red!36.20144865467842}{\strut ignorant} \colorbox{red!0.0}{\strut like} \colorbox{red!31.83824092889938}{\strut crack} \colorbox{red!31.83824092889938}{\strut \#\#er} \colorbox{red!0.0}{\strut mountain} \colorbox{red!17.62251573813954}{\strut hill} \colorbox{red!17.62251573813954}{\strut \#\#bill} \colorbox{red!17.62251573813954}{\strut \#\#ies}\endgroup} & HS \\
 
 & BERT-RP & {\begingroup\setlength{\fboxsep}{1pt} \colorbox{red!0.0}{\strut <user>} \colorbox{red!0.0}{\strut ff} \colorbox{red!0.0}{\strut \#\#s} \colorbox{red!0.0}{\strut these} \colorbox{red!45.35360285707457}{\strut ghetto} \colorbox{red!100.0}{\strut blacks} \colorbox{red!36.57737296812498}{\strut are} \colorbox{red!23.859449206708618}{\strut ignorant} \colorbox{red!0.0}{\strut like} \colorbox{red!26.30708855476805}{\strut crack} \colorbox{red!26.30708855476805}{\strut \#\#er} \colorbox{red!11.056221727591058}{\strut mountain} \colorbox{red!0.0}{\strut hill} \colorbox{red!0.0}{\strut \#\#bill} \colorbox{red!0.0}{\strut \#\#ies} \endgroup} & HS \\
  
 & BERT-HX & {\begingroup\setlength{\fboxsep}{1pt} \colorbox{red!0.0}{\strut <user>} \colorbox{red!0.0}{\strut ff} \colorbox{red!0.0}{\strut \#\#s} \colorbox{red!46.82091919644658}{\strut these} \colorbox{red!100.0}{\strut ghetto} \colorbox{red!0.0}{\strut blacks} \colorbox{red!0.0}{\strut are} \colorbox{red!32.16612634627618}{\strut ignorant} \colorbox{red!0.0}{\strut like} \colorbox{red!0.0}{\strut crack} \colorbox{red!0.0}{\strut \#\#er} \colorbox{red!0.0}{\strut mountain} \colorbox{red!50.588003439258}{\strut hill} \colorbox{red!50.588003439258}{\strut \#\#bill} \colorbox{red!50.588003439258}{\strut \#\#ies}   \endgroup} & HS \\
\hline
\hline

\hline
\multicolumn{4}{l}{\textbf{Case 2  Implicit hate speech}}\\
\hline

3 & Human & {\begingroup\setlength{\fboxsep}{0pt} \colorbox{red!0}{\strut yah} \colorbox{red!100.0}{\strut christian} \colorbox{red!100.0}{\strut island} \colorbox{red!100.0}{\strut republic} \colorbox{red!100.0}{\strut shall} \colorbox{red!100.0}{\strut not} \colorbox{red!100.0}{\strut have} \colorbox{red!100.0}{\strut queers} \colorbox{red!100.0}{\strut or} \colorbox{red!100.0}{\strut witches} \colorbox{red!100.0}{\strut on} \colorbox{red!100.0}{\strut it} \colorbox{red!100.0}{\strut if} \colorbox{red!100.0}{\strut we} \colorbox{red!100.0}{\strut find} \colorbox{red!100.0}{\strut any} \colorbox{red!100.0}{\strut death} \endgroup}  & HS \\
\hline

 & BERT-MRP & {\begingroup\setlength{\fboxsep}{0pt} \colorbox{red!0.0}{\strut ya} \colorbox{red!0.0}{\strut \#\#h} \colorbox{red!0.0}{\strut christian} \colorbox{red!0.0}{\strut island} \colorbox{red!0.0}{\strut republic} \colorbox{red!0.0}{\strut shall} \colorbox{red!0.0}{\strut not} \colorbox{red!69.02487896760483}{\strut have} \colorbox{red!100.0}{\strut queer} \colorbox{red!100.0}{\strut \#\#s} \colorbox{red!46.047030749677305}{\strut or} \colorbox{red!55.948289482026915}{\strut witches} \colorbox{red!0.0}{\strut on} \colorbox{red!0.0}{\strut it} \colorbox{red!0.0}{\strut if} \colorbox{red!0.0}{\strut we} \colorbox{red!47.87496479202237}{\strut find} \colorbox{red!0.0}{\strut any} \colorbox{red!71.70080886540703}{\strut death} \endgroup} & HS \\
 
 & BERT-RP & {\begingroup\setlength{\fboxsep}{0pt} \colorbox{red!0.0}{\strut ya} \colorbox{red!0.0}{\strut \#\#h} \colorbox{red!0.0}{\strut christian} \colorbox{red!0.0}{\strut island} \colorbox{red!0.0}{\strut republic} \colorbox{red!0.0}{\strut shall} \colorbox{red!0.0}{\strut not} \colorbox{red!54.99595278448809}{\strut have} \colorbox{red!100.0}{\strut queer} \colorbox{red!100.0}{\strut \#\#s} \colorbox{red!46.84191746645279}{\strut or} \colorbox{red!61.133440812592134}{\strut witches} \colorbox{red!35.04676177991256}{\strut on} \colorbox{red!0.0}{\strut it} \colorbox{red!0.0}{\strut if} \colorbox{red!0.0}{\strut we} \colorbox{red!0.0}{\strut find} \colorbox{red!0.0}{\strut any} \colorbox{red!35.495958683284314}{\strut death}\endgroup} & HS \\
 
  & BERT-HX & {\begingroup\setlength{\fboxsep}{0pt} \colorbox{red!0.0}{\strut ya} \colorbox{red!0.0}{\strut \#\#h} \colorbox{red!100.0}{\strut christian} \colorbox{red!0.0}{\strut island} \colorbox{red!0.0}{\strut republic} \colorbox{red!0.0}{\strut shall} \colorbox{red!0.0}{\strut not} \colorbox{red!0.0}{\strut have} \colorbox{red!0.0}{\strut queer} \colorbox{red!0.0}{\strut \#\#s} \colorbox{red!0.0}{\strut or} \colorbox{red!0.0}{\strut witches} \colorbox{red!0.0}{\strut on} \colorbox{red!0.0}{\strut it} \colorbox{red!0.0}{\strut if} \colorbox{red!0.0}{\strut we} \colorbox{red!0.0}{\strut find} \colorbox{red!0.0}{\strut any} \colorbox{red!0.0}{\strut death} \endgroup} & NO \\
  
\hline

4 &Human & {\begingroup\setlength{\fboxsep}{1pt} \colorbox{red!0.0}{\strut you} \colorbox{red!66.66666666666666}{\strut can} \colorbox{red!66.66666666666666}{\strut not} \colorbox{red!66.66666666666666}{\strut culturally} \colorbox{red!66.66666666666666}{\strut enrich} \colorbox{red!33.33333333333333}{\strut a} \colorbox{red!100.0}{\strut moslem} \endgroup} & HS \\ 
\hline
 
 & BERT-MRP & {\begingroup\setlength{\fboxsep}{1pt} \colorbox{red!10.867946167400689}{\strut you} \colorbox{red!0.0}{\strut can} \colorbox{red!0.0}{\strut not} \colorbox{red!0.0}{\strut culturally} \colorbox{red!6.009660084185607}{\strut en} \colorbox{red!6.009660084185607}{\strut \#\#rich} \colorbox{red!24.503327513887836}{\strut a} \colorbox{red!100.0}{\strut mo} \colorbox{red!100.0}{\strut \#\#sle} \colorbox{red!100.0}{\strut \#\#m} \endgroup} & HS \\
 
 & BERT-RP & {\begingroup\setlength{\fboxsep}{1pt} \colorbox{red!0.0}{\strut you} \colorbox{red!100.0}{\strut can} \colorbox{red!0.0}{\strut not} \colorbox{red!0.0}{\strut culturally} \colorbox{red!0.0}{\strut en} \colorbox{red!0.0}{\strut \#\#rich} \colorbox{red!0.0}{\strut a} \colorbox{red!0.0}{\strut mo} \colorbox{red!0.0}{\strut \#\#sle} \colorbox{red!0.0}{\strut \#\#m} \endgroup} & NO \\
 
  & BERT-HX & {\begingroup\setlength{\fboxsep}{1pt} \colorbox{red!0.0}{\strut you} \colorbox{red!0.0}{\strut can} \colorbox{red!0.0}{\strut not} \colorbox{red!0.0}{\strut culturally} \colorbox{red!0.0}{\strut en} \colorbox{red!0.0}{\strut \#\#rich} \colorbox{red!100.0}{\strut a} \colorbox{red!0.0}{\strut mo} \colorbox{red!0.0}{\strut \#\#sle} \colorbox{red!0.0}{\strut \#\#m} \endgroup} & NO \\
 
\hline

5 & Human & {\begingroup\setlength{\fboxsep}{1pt} \colorbox{red!0.0}{\strut <user>} \colorbox{red!0.0}{\strut he} \colorbox{red!0.0}{\strut is} \colorbox{red!0.0}{\strut infected} \colorbox{red!0.0}{\strut with} \colorbox{red!100.0}{\strut jihadi} \colorbox{red!66.66666666666666}{\strut virus} \colorbox{red!33.33333333333333}{\strut he} \colorbox{red!33.33333333333333}{\strut will} \colorbox{red!33.33333333333333}{\strut spread} \colorbox{red!33.33333333333333}{\strut it} \colorbox{red!33.33333333333333}{\strut to} \colorbox{red!33.33333333333333}{\strut others}\endgroup} & HS \\
\hline

 & BERT-MRP & {\begingroup\setlength{\fboxsep}{1pt} \colorbox{red!0.0}{\strut <user>} \colorbox{red!42.17458041073887}{\strut he} \colorbox{red!0.0}{\strut is} \colorbox{red!0.0}{\strut infected} \colorbox{red!0.0}{\strut with} \colorbox{red!100.0}{\strut jihad} \colorbox{red!100.0}{\strut \#\#i} \colorbox{red!0.0}{\strut virus} \colorbox{red!36.75097580899456}{\strut he} \colorbox{red!28.40903630858377}{\strut will} \colorbox{red!26.128269097819533}{\strut spread} \colorbox{red!0.0}{\strut it} \colorbox{red!0.0}{\strut to} \colorbox{red!27.97481080090254}{\strut others}\endgroup} & HS \\
 
 & BERT-RP & {\begingroup\setlength{\fboxsep}{1pt} \colorbox{red!21.716832591510666}{\strut <user>} \colorbox{red!39.3313376486578}{\strut he} \colorbox{red!0.0}{\strut is} \colorbox{red!51.79633468331034}{\strut infected} \colorbox{red!0.0}{\strut with} \colorbox{red!100.0}{\strut jihad} \colorbox{red!100.0}{\strut \#\#i} \colorbox{red!0.0}{\strut virus} \colorbox{red!31.2525693608857}{\strut he} \colorbox{red!0.0}{\strut will} \colorbox{red!0.0}{\strut spread} \colorbox{red!0.0}{\strut it} \colorbox{red!24.01862270550423}{\strut to} \colorbox{red!0.0}{\strut others} \endgroup} & HS  \\
 
 & BERT-HX & {\begingroup\setlength{\fboxsep}{1pt} \colorbox{red!0.0}{\strut <user>} \colorbox{red!0.0}{\strut he} \colorbox{red!0.0}{\strut is} \colorbox{red!7.275644981557754}{\strut infected} \colorbox{red!10.004349469203785}{\strut with} \colorbox{red!100.0}{\strut jihad} \colorbox{red!100.0}{\strut \#\#i} \colorbox{red!7.908027088468128}{\strut virus} \colorbox{red!0.0}{\strut he} \colorbox{red!0.0}{\strut will} \colorbox{red!6.946051670768738}{\strut spread} \colorbox{red!7.610364187259748}{\strut it} \colorbox{red!0.0}{\strut to} \colorbox{red!0.0}{\strut others} \endgroup} & OF \\
\hline

\end{tabular}
\caption{The highlighted words of the human rationale and the rationale of the models with detection results. BERT-HX is BERT-HateXplain. In the label column, the ground truth is of humans and the remaining labels are the predictions of each model. HS is 'hate speech,' OF is 'offensive,' and NO is 'normal.' More examples are in Appendix.}
\label{attention_vis_2}
}

\end{table*}

\noindent
\textbf{Bias-based metrics}  For the model bias, the proposed models show superior results compared to other models. According to Table~\ref{perform_n_bias}, the models trained using the rationales achieve higher scores than others in general. Given that the human rationales of hatred imply that hate speech is judged based on the context, not merely specific expressions, learning the rationale can exclude the model bias towards the particular words for the prediction. Meanwhile, the proposed BERT-RP and BERT-MRP show better performance than BERT-HateXplain, even though they all utilize the rationale in training. BERT-MRP shows the best scores, and BERT-RP is the second-best for all three metrics. Additionally, Figure~\ref{bias_graph_sub} shows the scores of the models for each of the ten major target groups. It can be seen that the proposed models score evenly high for all the target groups. While other models have significant differences in their bias depending on the groups, the proposed models have comparatively no correlation with them. 

\noindent
\textbf{Explainability-based metrics}  In terms of explainability, the proposed models still perform better than others overall. For Plausibility, BERT-MRP achieves the best performance for all three metrics. It scores much higher than others because it is allowed to directly guess the human rationales during the intermediate training stage. For Faithfulness, BERT-HateXplain[LIME] shows the highest score for Comprehensiveness, and BERT-MLM[LIME] is the best for Sufficiency. However, these models do not reliably score well when considering all four scores obtained according to each of the two measurement methods: attention values or LIME. They show worse scores than BERT for the rest of the scores. On the other hand, BERT-MRP and BERT-RP offer stably high performance for all five explainability-based metrics. 

Based on all these results, BERT-MRP and BERT-RP demonstrate the best performance overall for the three types of metrics. Thus, learning the human rationale as an intermediate task before training on hate speech detection seems effective for detection performance and model bias and its explainability. This framework contributes to better performance than the other--pre-finetuning on MLM as well as another way of using rationales. 

BERT-MRP generally achieves better results than BERT-RP, wherein the intermediate task is basically the same as token classification. 
The plots in Figure~\ref{ratio_graph} show the change in test scores according to the masking ratio in MRP. According to Figure~\ref{ratio_graph}(a), when more than 20\% of all rationales were masked, there is no significant difference in the token classification performance, although the amount of loss decreases as the ratio decreases. 
When each model was re-trained for hate speech detection, as shown in Figure~\ref{ratio_graph}(b), the case of 50\% ratio in BERT-MRP achieved the best classification performance. As MRP is a method for inferring the rationale of a particular token based on surrounding tokens, the model can successfully learn the human rationale within the context. Learning parameters during this reasoning process based on context seems to effectively prevent biased prediction while still being explainable and consequently improves the detection performance substantially. 

\subsection{Qualitative results}
Table~\ref{attention_vis_2} shows examples of detection results from models that use human rationale for their training. The visualized values as the model rationales are the LIME results used to measure the explainability-based scores. For the human ground truth, the average value per word of human-annotated rationales is expressed for each word. The darker the color, the higher the values. 

It is relatively easy to judge explicit hate speech that includes clear derogatory expressions. As shown in Case 1 of Table~\ref{attention_vis_2}, all the models perform well. The human rationale tends to focus on specific abusive words, and so does the rationale of each model. However, the rationales of the two proposed models match the ground truth better than BERT-HateXplain. Our method to train a model on a token classification-based task leads well the model to focus on human-like grounds in the sentence by directly learning the human rationale. 

As in Case 2, the implicit hate speech with no aggressive expressions cannot be grasped through context. The human rationale thus tends to appear throughout the sentence. As this might make hate speech detection relatively challenging, the detection results of BERT-HateXplain or BERT-RP seem incorrect for some sentences. However, BERT-MRP works accurately based on its rationale that is much more similar to human's than others. Meanwhile, BERT-HateXplain shows a low matching rate of the rationale when the human rationale is throughout the sentence. It uses the human rationale as the ground truth attention, and if there is no difference in the human rationale across tokens, the ground truth could become similar to that of any normal sentence represented by uniform values. This affects the model's explainability and may lead to incorrect detection results. The proposed method does not cause that problem. It gets the distinguishable ground truth from normal ones and assigns it as labels to tokens.  
On the other hand, the rationale of BERT-MRP matches the ground truth better than that of BERT-RP. MRP requires more context-awareness ability when predicting the masked token by allowing the model to consider the abusiveness of surrounding words that are provided corresponding human rationale. This offers robust detection performance, even when it is necessary to understand the context.\\
\vspace{-0.3cm}
\section{Conclusion}
This paper presents a method to implement a hate speech detection model considering bias and explainability. We adopt a framework to finetune a pre-trained language model in two stages. As the intermediate task, we propose Masked Rationale Prediction (MRP), which predicts masked rationales for some tokens with the additional rationale information of the remaining surrounding tokens. With this, the model learns to identify abusiveness for each token and the human reasoning process based on context. The trained model by MRP is finetuned again on hate speech detection.  

As a result, across quantitative and qualitative evaluations, the proposed model shows state-of-the-art performance in bias and explainability, as well as the detection result. 
And the examples demonstrate its robustness in detecting hate speech, whether explicit or implicit, based on its superior explainability. Meanwhile, we experimented with only BERT as the pre-trained model to compare our method with base models. But any other Transformer encoder-based model can be easily applied, which can be taken as future work. \\

\section*{Acknowledgements}
This work was supported by the Ministry of Education of the Republic of Korea and the National Research Foundation of Korea (NRF) (NRF-2021S1A5A2A03065899), and also by the NRF of Korea grant funded by the Korean government (MSIT) (NRF-2022R1A2C1007434). \\

\bibliography{acl.bbl}
\bibliographystyle{acl_natbib}

\clearpage
\setcounter{table}{0}
\renewcommand{\thetable}{A\arabic{table}}
\setcounter{figure}{0}
\renewcommand{\thefigure}{A\arabic{figure}}
\pagenumbering{arabic}
\renewcommand*{\thepage}{A\arabic{page}}

\appendix

\label{sec:appendix} 

\vspace{-4cm}

\begin{figure*}[!b]
\centering 
\includegraphics[width=0.95\linewidth]{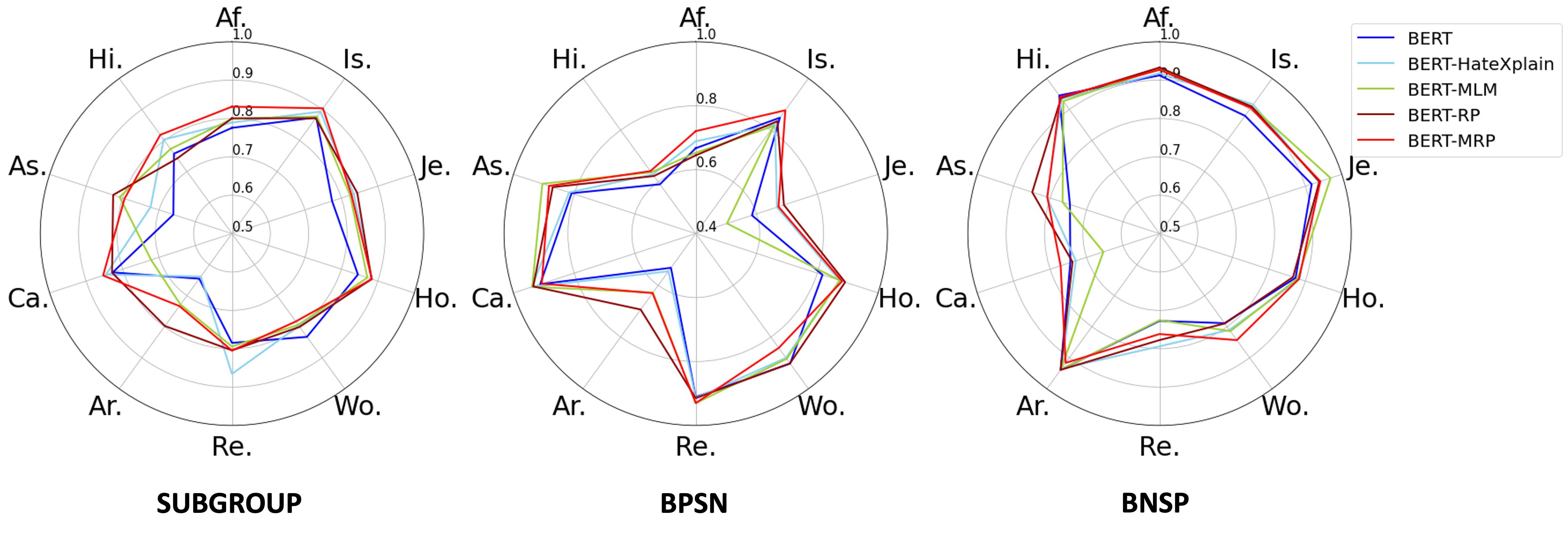}
\caption{The Bias-based scores for each 10 target groups. The target group labels are `African', `Islam', `Jewish',
`Homosexual', `Women', `Refugee', `Arab', `Caucasian', `Asian', and `Hispanic' in clockwise direction respectively.}
\label{bias_graph_all}
\end{figure*}

\vspace{4cm}


\begin{table*}[!b]
\setstretch{1.2}
\renewcommand{\arraystretch}{1.3}
\renewcommand{\tabcolsep}{0.7mm}
\centering
{\small
\begin{tabular}{lllll}
\hline
\textbf{Model} & \textbf{Example sentence and Rationale} & \textbf{Label} \\
\hline
\hline

\hline

Human & {\begingroup\setlength{\fboxsep}{1pt}
\colorbox{red!0.0}{\strut if} \colorbox{red!0.0}{\strut you} \colorbox{red!0.0}{\strut ’} \colorbox{red!0.0}{\strut re} \colorbox{red!0.0}{\strut an} \colorbox{red!0.0}{\strut army} \colorbox{red!0.0}{\strut that} \colorbox{red!0.0}{\strut is} \colorbox{red!0.0}{\strut republican} \colorbox{red!0.0}{\strut you} \colorbox{red!0.0}{\strut are} \colorbox{red!0.0}{\strut not} \colorbox{red!0.0}{\strut welcome} \colorbox{red!0.0}{\strut here} \colorbox{red!50.0}{\strut we} \colorbox{red!100.0}{\strut hate} \colorbox{red!100.0}{\strut you} \endgroup} & HS \\ 
\hline
 
BERT-MRP & {\begingroup\setlength{\fboxsep}{1pt}  \colorbox{red!0.0}{\strut if} \colorbox{red!0.0}{\strut you} \colorbox{red!0.0}{\strut ’} \colorbox{red!0.0}{\strut re} \colorbox{red!0.0}{\strut an} \colorbox{red!0.0}{\strut army} \colorbox{red!39.34601896963705}{\strut that} \colorbox{red!37.27061298395418}{\strut is} \colorbox{red!0.0}{\strut republican} \colorbox{red!0.0}{\strut you} \colorbox{red!0.0}{\strut are} \colorbox{red!0.0}{\strut not} \colorbox{red!57.82033155300345}{\strut welcome} \colorbox{red!75.37462598288242}{\strut here} \colorbox{red!0.0}{\strut we} \colorbox{red!100.0}{\strut hate} \colorbox{red!66.30245822828313}{\strut you} \endgroup} & HS \\
BERT-RP & {\begingroup\setlength{\fboxsep}{1pt}  \colorbox{red!0}{\strut if} \colorbox{red!0}{\strut you} \colorbox{red!0}{\strut ’} \colorbox{red!0}{\strut re} \colorbox{red!0}{\strut an} \colorbox{red!0}{\strut army} \colorbox{red!0}{\strut that} \colorbox{red!0}{\strut is} \colorbox{red!0}{\strut republican} \colorbox{red!0}{\strut you} \colorbox{red!0}{\strut are} \colorbox{red!0}{\strut not} \colorbox{red!0}{\strut welcome} \colorbox{red!0}{\strut here} \colorbox{red!0}{\strut we} \colorbox{red!0}{\strut hate} \colorbox{red!0}{\strut you}\endgroup} & NO \\
BERT-HX & {\begingroup\setlength{\fboxsep}{1pt} \colorbox{red!0}{\strut if} \colorbox{red!0}{\strut you} \colorbox{red!0}{\strut ’} \colorbox{red!0}{\strut re} \colorbox{red!0}{\strut an} \colorbox{red!0}{\strut army} \colorbox{red!0}{\strut that} \colorbox{red!0}{\strut is} \colorbox{red!0}{\strut republican} \colorbox{red!0}{\strut you} \colorbox{red!0}{\strut are} \colorbox{red!0}{\strut not} \colorbox{red!0}{\strut welcome} \colorbox{red!0}{\strut here} \colorbox{red!0}{\strut we} \colorbox{red!0}{\strut hate} \colorbox{red!0}{\strut you}\endgroup} & NO \\
\hline

Human & {\begingroup\setlength{\fboxsep}{0pt} \colorbox{red!0.0}{\strut the} \colorbox{red!0.0}{\strut german} \colorbox{red!0.0}{\strut regime} \colorbox{red!0.0}{\strut is} \colorbox{red!0.0}{\strut more} \colorbox{red!0.0}{\strut scared} \colorbox{red!0.0}{\strut of} \colorbox{red!0.0}{\strut women} \colorbox{red!0.0}{\strut like} \colorbox{red!0.0}{\strut this} \colorbox{red!0.0}{\strut than} \colorbox{red!0.0}{\strut than} \colorbox{red!0.0}{\strut are} \colorbox{red!0.0}{\strut of} \colorbox{red!0.0}{\strut any} \colorbox{red!100.0}{\strut moslem}
\colorbox{red!100.0}{\strut terrorist} \endgroup} & HS \\
\hline
BERT-MRP & {\begingroup\setlength{\fboxsep}{0pt} \colorbox{red!0.0}{\strut the} \colorbox{red!0.0}{\strut german} \colorbox{red!0.0}{\strut regime} \colorbox{red!0.0}{\strut is} \colorbox{red!0.0}{\strut more} \colorbox{red!19.09115439545584}{\strut scared} \colorbox{red!16.18210614683594}{\strut of} \colorbox{red!17.38104928378045}{\strut women} \colorbox{red!0.0}{\strut like} \colorbox{red!0.0}{\strut this} \colorbox{red!0.0}{\strut than} \colorbox{red!0.0}{\strut than} \colorbox{red!0.0}{\strut are} \colorbox{red!0.0}{\strut of} \colorbox{red!24.72568853424022}{\strut any} \colorbox{red!100.0}{\strut mo} \colorbox{red!100.0}{\strut \#\#sle} \colorbox{red!100.0}{\strut \#\#m} \colorbox{red!38.505119818849515}{\strut terrorist} \endgroup} & HS \\
BERT-RP & {\begingroup\setlength{\fboxsep}{0pt} \colorbox{red!0.0}{\strut the} \colorbox{red!0.0}{\strut german} \colorbox{red!14.593425019125474}{\strut regime} \colorbox{red!0.0}{\strut is} \colorbox{red!0.0}{\strut more} \colorbox{red!21.383858676084582}{\strut scared} \colorbox{red!17.508882474887884}{\strut of} \colorbox{red!19.907845161880307}{\strut women} \colorbox{red!0.0}{\strut like} \colorbox{red!0.0}{\strut this} \colorbox{red!0.0}{\strut than} \colorbox{red!0.0}{\strut than} \colorbox{red!0.0}{\strut are} \colorbox{red!0.0}{\strut of} \colorbox{red!0.0}{\strut any} \colorbox{red!100.0}{\strut mo} \colorbox{red!100.0}{\strut \#\#sle} \colorbox{red!100.0}{\strut \#\#m} \colorbox{red!56.256545360399855}{\strut terrorist}  \endgroup} & HS \\
BERT-HX & {\begingroup\setlength{\fboxsep}{0pt} \colorbox{red!0}{\strut the} \colorbox{red!0}{\strut german} \colorbox{red!0}{\strut regime} \colorbox{red!0}{\strut is} \colorbox{red!0}{\strut more} \colorbox{red!0}{\strut scared} \colorbox{red!0}{\strut of} \colorbox{red!0}{\strut women} \colorbox{red!0}{\strut like} \colorbox{red!0}{\strut this} \colorbox{red!0}{\strut than} \colorbox{red!0}{\strut than} \colorbox{red!0}{\strut are} \colorbox{red!0}{\strut of} \colorbox{red!0}{\strut any} \colorbox{red!0}{\strut mo} \colorbox{red!0}{\strut \#\#sle} \colorbox{red!0}{\strut \#\#m} \colorbox{red!0}{\strut terrorist}  \endgroup} & NO \\
\hline

Human & {\begingroup\setlength{\fboxsep}{1pt}
\colorbox{red!0.0}{\strut i} \colorbox{red!0.0}{\strut poisoned} \colorbox{red!0.0}{\strut sergei} \colorbox{red!0.0}{\strut skripal} \colorbox{red!0.0}{\strut he} \colorbox{red!0.0}{\strut was} \colorbox{red!0.0}{\strut a} \colorbox{red!100.0}{\strut faggot} \colorbox{red!0.0}{\strut anyway} \endgroup} & HS \\ 
\hline
BERT-MRP & {\begingroup\setlength{\fboxsep}{1pt}  \colorbox{red!9.036899060684386}{\strut i} \colorbox{red!0.0}{\strut poisoned} \colorbox{red!0.0}{\strut sergei} \colorbox{red!9.561660283815293}{\strut sk} \colorbox{red!9.561660283815293}{\strut \#\#rip} \colorbox{red!9.561660283815293}{\strut \#\#al} \colorbox{red!0.0}{\strut he} \colorbox{red!0.0}{\strut was} \colorbox{red!20.461555717080476}{\strut a} \colorbox{red!100.0}{\strut fa} \colorbox{red!100.0}{\strut \#\#gg} \colorbox{red!100.0}{\strut \#\#ot} \colorbox{red!0.0}{\strut anyway} \endgroup} & HS \\ 

BERT-RP & {\begingroup\setlength{\fboxsep}{1pt} \colorbox{red!11.164193638096082}{\strut i} \colorbox{red!0.0}{\strut poisoned} \colorbox{red!0.0}{\strut sergei} \colorbox{red!5.675591184488672}{\strut sk} \colorbox{red!5.675591184488672}{\strut \#\#rip} \colorbox{red!5.675591184488672}{\strut \#\#al} \colorbox{red!0.0}{\strut he} \colorbox{red!0.0}{\strut was} \colorbox{red!10.27689110257881}{\strut a} \colorbox{red!100.0}{\strut fa} \colorbox{red!100.0}{\strut \#\#gg} \colorbox{red!100.0}{\strut \#\#ot} \colorbox{red!0.0}{\strut anyway} \endgroup} & HS \\ 

BERT-HX & {\begingroup\setlength{\fboxsep}{1pt} \colorbox{red!0.0}{\strut i} \colorbox{red!0.0}{\strut poisoned} \colorbox{red!23.23484223553531}{\strut sergei} \colorbox{red!0.0}{\strut sk} \colorbox{red!0.0}{\strut \#\#rip} \colorbox{red!0.0}{\strut \#\#al} \colorbox{red!22.69525729064024}{\strut he} \colorbox{red!15.039507932980655}{\strut was} \colorbox{red!0.0}{\strut a} \colorbox{red!100.0}{\strut fa} \colorbox{red!100.0}{\strut \#\#gg} \colorbox{red!100.0}{\strut \#\#ot} \colorbox{red!15.775722360688924}{\strut anyway}  \endgroup} & OF \\ 
\hline


Human & {\begingroup\setlength{\fboxsep}{1pt} \colorbox{red!0.0}{\strut <user>} \colorbox{red!0.0}{\strut would} \colorbox{red!0.0}{\strut have} \colorbox{red!0.0}{\strut pressed} \colorbox{red!0.0}{\strut that} \colorbox{red!100.0}{\strut bitch} \colorbox{red!33.33333333333333}{\strut and} \colorbox{red!33.33333333333333}{\strut beat} \colorbox{red!33.33333333333333}{\strut the} \colorbox{red!100.0}{\strut faggot} \colorbox{red!33.33333333333333}{\strut out} \colorbox{red!33.33333333333333}{\strut of} \colorbox{red!33.33333333333333}{\strut him}  \endgroup} & HS \\ 
\hline
BERT-MRP & {\begingroup\setlength{\fboxsep}{1pt} \colorbox{red!0.0}{\strut <user>} \colorbox{red!21.81468710742534}{\strut would} \colorbox{red!0.0}{\strut have} \colorbox{red!0.0}{\strut pressed} \colorbox{red!0.0}{\strut that} \colorbox{red!0.0}{\strut bitch} \colorbox{red!17.086766315957355}{\strut and} \colorbox{red!24.151424839476217}{\strut beat} \colorbox{red!56.45253834631223}{\strut the} \colorbox{red!100.0}{\strut fa} \colorbox{red!100.0}{\strut \#\#gg} \colorbox{red!100.0}{\strut \#\#ot} \colorbox{red!0.0}{\strut out} \colorbox{red!0.0}{\strut of} \colorbox{red!0.0}{\strut him} \endgroup} & HS \\ 

BERT-RP & {\begingroup\setlength{\fboxsep}{1pt} \colorbox{red!33.358068406519465}{\strut <user>} \colorbox{red!0.0}{\strut would} \colorbox{red!0.0}{\strut have} \colorbox{red!0.0}{\strut pressed} \colorbox{red!27.04978257043416}{\strut that} \colorbox{red!100.0}{\strut bitch} \colorbox{red!0.0}{\strut and} \colorbox{red!17.36013671390747}{\strut beat} \colorbox{red!0.0}{\strut the} \colorbox{red!45.44575108271631}{\strut fa} \colorbox{red!45.44575108271631}{\strut \#\#gg} \colorbox{red!45.44575108271631}{\strut \#\#ot} \colorbox{red!0.0}{\strut out} \colorbox{red!0.0}{\strut of} \colorbox{red!0.0}{\strut him}   \endgroup} & HS \\ 

BERT-HX & {\begingroup\setlength{\fboxsep}{1pt} \colorbox{red!0.0}{\strut <user>} \colorbox{red!15.909313688600454}{\strut would} \colorbox{red!0.0}{\strut have} \colorbox{red!0.0}{\strut pressed} \colorbox{red!0.0}{\strut that} \colorbox{red!0.0}{\strut bitch} \colorbox{red!0.0}{\strut and} \colorbox{red!12.200695820990799}{\strut beat} \colorbox{red!21.55664725340973}{\strut the} \colorbox{red!100.0}{\strut fa} \colorbox{red!100.0}{\strut \#\#gg} \colorbox{red!100.0}{\strut \#\#ot} \colorbox{red!15.069318615129527}{\strut out} \colorbox{red!10.235288419073399}{\strut of} \colorbox{red!0.0}{\strut him}  \endgroup} & OF \\ 
\hline

Human & {\begingroup\setlength{\fboxsep}{1pt} \colorbox{red!0.0}{\strut massive} \colorbox{red!100.0}{\strut muzrat} \colorbox{red!0.0}{\strut espionage} \endgroup} & HS \\
\hline
BERT-MRP & {\begingroup\setlength{\fboxsep}{1pt} \colorbox{red!4.043632546332093}{\strut massive} \colorbox{red!100.0}{\strut mu} \colorbox{red!100.0}{\strut \#\#z} \colorbox{red!100.0}{\strut \#\#rat} \colorbox{red!0.0}{\strut espionage} \endgroup} & HS \\
BERT-RP & {\begingroup\setlength{\fboxsep}{1pt} \colorbox{red!8.476154810854595}{\strut massive} \colorbox{red!100.0}{\strut mu} \colorbox{red!100.0}{\strut \#\#z} \colorbox{red!100.0}{\strut \#\#rat} \colorbox{red!0.0}{\strut espionage} \endgroup} & HS \\
BERT-HX & {\begingroup\setlength{\fboxsep}{1pt} \colorbox{red!16.01308095773438}{\strut massive} \colorbox{red!100.0}{\strut mu} \colorbox{red!100.0}{\strut \#\#z} \colorbox{red!100.0}{\strut \#\#rat} \colorbox{red!0.0}{\strut espionage} \endgroup} & HS \\
\hline

Human & {\begingroup\setlength{\fboxsep}{1pt} \colorbox{red!0.0}{\strut enjoy} \colorbox{red!0.0}{\strut your} \colorbox{red!100.0}{\strut kike} \colorbox{red!0.0}{\strut faith}   \endgroup} & HS \\ 
\hline
BERT-MRP & {\begingroup\setlength{\fboxsep}{1pt} \colorbox{red!0.0}{\strut enjoy} \colorbox{red!0.0}{\strut your} \colorbox{red!100.0}{\strut ki} \colorbox{red!100.0}{\strut \#\#ke} \colorbox{red!0.0}{\strut faith}   \endgroup}& HS\\
BERT-RP & {\begingroup\setlength{\fboxsep}{1pt} \colorbox{red!0.0}{\strut enjoy} \colorbox{red!0.0}{\strut your} \colorbox{red!100.0}{\strut ki} \colorbox{red!100.0}{\strut \#\#ke} \colorbox{red!0.0}{\strut faith}   \endgroup} & HS \\
BERT-HX & {\begingroup\setlength{\fboxsep}{1pt}  \colorbox{red!0.0}{\strut enjoy} \colorbox{red!0.0}{\strut your} \colorbox{red!100.0}{\strut ki} \colorbox{red!100.0}{\strut \#\#ke} \colorbox{red!0.0}{\strut faith}   \endgroup} & HS \\
\hline

\hline

\end{tabular}
\caption{The highlighted words of the human rationale and the rationale of the models with detection results in 'hate speech' sentences. }
\label{hs_appendix}
}

\end{table*}


\begin{table*}[!b]
\setstretch{1.2}
\renewcommand{\arraystretch}{1.3}
\renewcommand{\tabcolsep}{0.7mm}
\centering
{\small
\begin{tabular}{lllll}
\hline
\textbf{Model} & \textbf{Example sentence and Rationale} & \textbf{Label} \\
\hline
\hline

Human & {\begingroup\setlength{\fboxsep}{1pt} \colorbox{red!0.0}{\strut “} \colorbox{red!0.0}{\strut i} \colorbox{red!66.66666666666666}{\strut don} \colorbox{red!66.66666666666666}{\strut ’} \colorbox{red!66.66666666666666}{\strut t} \colorbox{red!66.66666666666666}{\strut have} \colorbox{red!66.66666666666666}{\strut to} \colorbox{red!66.66666666666666}{\strut respect} \colorbox{red!66.66666666666666}{\strut women} \colorbox{red!66.66666666666666}{\strut i} \colorbox{red!66.66666666666666}{\strut ’} \colorbox{red!66.66666666666666}{\strut m} \colorbox{red!66.66666666666666}{\strut gay} \colorbox{red!33.33333333333333}{\strut ”} \colorbox{red!0.0}{\strut kevin}  \endgroup} & OF\\ 
\hline
BERT-MRP & {\begingroup\setlength{\fboxsep}{1pt}  \colorbox{red!47}{\strut “} \colorbox{red!17}{\strut i} \colorbox{red!6}{\strut don} \colorbox{red!0}{\strut ’} \colorbox{red!4}{\strut t} \colorbox{red!5}{\strut have} \colorbox{red!3}{\strut to} \colorbox{red!59}{\strut respect} \colorbox{red!37}{\strut women} \colorbox{red!34}{\strut i} \colorbox{red!3}{\strut ’} \colorbox{red!25}{\strut m} \colorbox{red!100}{\strut gay} \colorbox{red!23}{\strut ”} \colorbox{red!65}{\strut kevin}   \endgroup} & OF\\
BERT-RP & {\begingroup\setlength{\fboxsep}{1pt}  \colorbox{red!20}{\strut “} \colorbox{red!6}{\strut i} \colorbox{red!0}{\strut don} \colorbox{red!61}{\strut ’} \colorbox{red!1}{\strut t} \colorbox{red!0}{\strut have} \colorbox{red!0}{\strut to} \colorbox{red!25}{\strut respect} \colorbox{red!31}{\strut women} \colorbox{red!45}{\strut i} \colorbox{red!9}{\strut ’} \colorbox{red!45}{\strut m} \colorbox{red!100}{\strut gay} \colorbox{red!26}{\strut ”} \colorbox{red!43}{\strut kevin}    \endgroup} & NO \\
BERT-HX & {\begingroup\setlength{\fboxsep}{1pt}  \colorbox{red!0}{\strut “} \colorbox{red!0}{\strut i} \colorbox{red!0}{\strut don} \colorbox{red!0}{\strut ’} \colorbox{red!0}{\strut t} \colorbox{red!0}{\strut have} \colorbox{red!0}{\strut to} \colorbox{red!0}{\strut respect} \colorbox{red!0}{\strut women} \colorbox{red!0}{\strut i} \colorbox{red!0}{\strut ’} \colorbox{red!0}{\strut m} \colorbox{red!0}{\strut gay} \colorbox{red!0}{\strut ”} \colorbox{red!100}{\strut kevin}  \endgroup} & NO \\
\hline

Human & {\begingroup\setlength{\fboxsep}{1pt} \colorbox{red!0.0}{\strut logan} \colorbox{red!33.33333333333333}{\strut paul} \colorbox{red!33.33333333333333}{\strut is} \colorbox{red!33.33333333333333}{\strut a} \colorbox{red!66.66666666666666}{\strut fucking} \colorbox{red!66.66666666666666}{\strut bozo}   \endgroup} & OF \\ 
\hline
BERT-MRP & {\begingroup\setlength{\fboxsep}{1pt} \colorbox{red!12}{\strut logan} \colorbox{red!17}{\strut paul} \colorbox{red!0}{\strut is} \colorbox{red!2}{\strut a} \colorbox{red!11}{\strut fucking} \colorbox{red!69}{\strut bo} \colorbox{red!100}{\strut \#\#zo}   \endgroup} & OF\\
BERT-RP & {\begingroup\setlength{\fboxsep}{1pt}  \colorbox{red!12}{\strut logan} \colorbox{red!0}{\strut paul} \colorbox{red!1}{\strut is} \colorbox{red!0}{\strut a} \colorbox{red!81}{\strut fucking} \colorbox{red!100}{\strut bo} \colorbox{red!86}{\strut \#\#zo}    \endgroup} & NO \\
BERT-HX & {\begingroup\setlength{\fboxsep}{1pt}  \colorbox{red!100}{\strut logan} \colorbox{red!0}{\strut paul} \colorbox{red!26}{\strut is} \colorbox{red!38}{\strut a} \colorbox{red!0}{\strut fucking} \colorbox{red!0}{\strut bo} \colorbox{red!0}{\strut \#\#zo}   \endgroup} & NO \\
\hline

Human & {\begingroup\setlength{\fboxsep}{1pt} \colorbox{red!0.0}{\strut <user>} \colorbox{red!0.0}{\strut <user>} \colorbox{red!0.0}{\strut there} \colorbox{red!0.0}{\strut are} \colorbox{red!0.0}{\strut literally} \colorbox{red!33.33333333333333}{\strut nazis} \colorbox{red!0.0}{\strut all} \colorbox{red!0.0}{\strut over} \colorbox{red!0.0}{\strut all} \colorbox{red!0.0}{\strut the} \colorbox{red!0.0}{\strut time} \colorbox{red!0.0}{\strut rigjt} \colorbox{red!0.0}{\strut now} \colorbox{red!33.33333333333333}{\strut throw} \colorbox{red!33.33333333333333}{\strut a} \colorbox{red!33.33333333333333}{\strut rock}  \endgroup} & OF \\
 & {\begingroup\setlength{\fboxsep}{1pt}
\colorbox{red!33.33333333333333}{\strut you} \colorbox{red!33.33333333333333}{\strut hit} \colorbox{red!33.33333333333333}{\strut a} \colorbox{red!66.66666666666666}{\strut nazi}   \endgroup} & \\ 
\hline
BERT-MRP & {\begingroup\setlength{\fboxsep}{1pt}  \colorbox{red!12}{\strut <user>} \colorbox{red!13}{\strut <user>} \colorbox{red!8}{\strut there} \colorbox{red!16}{\strut are} \colorbox{red!44}{\strut literally} \colorbox{red!71}{\strut nazis} \colorbox{red!5}{\strut all} \colorbox{red!5}{\strut over} \colorbox{red!3}{\strut all} \colorbox{red!0}{\strut the} \colorbox{red!1}{\strut time} \colorbox{red!1}{\strut rig} \colorbox{red!4}{\strut \#\#j} \colorbox{red!0}{\strut \#\#t} \colorbox{red!2}{\strut now} \colorbox{red!8}{\strut throw} \colorbox{red!0}{\strut a} \colorbox{red!30}{\strut rock} \endgroup} & OF \\
 & {\begingroup\setlength{\fboxsep}{1pt}   
 \colorbox{red!10}{\strut you} \colorbox{red!36}{\strut hit} \colorbox{red!15}{\strut a} \colorbox{red!100}{\strut nazi}   \endgroup} & \\
BERT-RP & {\begingroup\setlength{\fboxsep}{1pt}   \colorbox{red!11}{\strut <user>} \colorbox{red!19}{\strut <user>} \colorbox{red!2}{\strut there} \colorbox{red!5}{\strut are} \colorbox{red!18}{\strut literally} \colorbox{red!46}{\strut nazis} \colorbox{red!9}{\strut all} \colorbox{red!6}{\strut over} \colorbox{red!3}{\strut all} \colorbox{red!0}{\strut the} \colorbox{red!3}{\strut time} \colorbox{red!2}{\strut rig} \colorbox{red!2}{\strut \#\#j} \colorbox{red!3}{\strut \#\#t} \colorbox{red!2}{\strut now} \colorbox{red!1}{\strut throw} \colorbox{red!0}{\strut a} \colorbox{red!14}{\strut rock}   \endgroup} & OF \\
 & {\begingroup\setlength{\fboxsep}{1pt}
 \colorbox{red!3}{\strut you} \colorbox{red!44}{\strut hit} \colorbox{red!42}{\strut a} \colorbox{red!100}{\strut nazi}   \endgroup} &  \\
BERT-HX & {\begingroup\setlength{\fboxsep}{1pt}  \colorbox{red!100}{\strut <user>} \colorbox{red!0}{\strut <user>} \colorbox{red!0}{\strut there} \colorbox{red!0}{\strut are} \colorbox{red!0}{\strut literally} \colorbox{red!0}{\strut nazis} \colorbox{red!0}{\strut all} \colorbox{red!0}{\strut over} \colorbox{red!0}{\strut all} \colorbox{red!0}{\strut the} \colorbox{red!0}{\strut time} \colorbox{red!0}{\strut rig} \colorbox{red!0}{\strut \#\#j} \colorbox{red!0}{\strut \#\#t} \colorbox{red!0}{\strut now} \colorbox{red!0}{\strut throw} \colorbox{red!0}{\strut a} \colorbox{red!0}{\strut rock} \endgroup} & NO \\
 & {\begingroup\setlength{\fboxsep}{1pt}
 \colorbox{red!0}{\strut you} \colorbox{red!0}{\strut hit} \colorbox{red!66}{\strut a} \colorbox{red!0}{\strut nazi}   \endgroup} & \\
\hline

Human & {\begingroup\setlength{\fboxsep}{1pt} \colorbox{red!33.33333333333333}{\strut all} \colorbox{red!33.33333333333333}{\strut my} \colorbox{red!33.33333333333333}{\strut friends} \colorbox{red!33.33333333333333}{\strut and} \colorbox{red!33.33333333333333}{\strut peers} \colorbox{red!33.33333333333333}{\strut are} \colorbox{red!33.33333333333333}{\strut being} \colorbox{red!33.33333333333333}{\strut openly} \colorbox{red!66.66666666666666}{\strut racist} \colorbox{red!66.66666666666666}{\strut towards} \colorbox{red!66.66666666666666}{\strut asians} \colorbox{red!33.33333333333333}{\strut and} \colorbox{red!33.33333333333333}{\strut this} \colorbox{red!66.66666666666666}{\strut bitch} \colorbox{red!33.33333333333333}{\strut is} 
\endgroup} & OF \\
 & {\begingroup\setlength{\fboxsep}{1pt}
\colorbox{red!33.33333333333333}{\strut not} \colorbox{red!33.33333333333333}{\strut having} \colorbox{red!33.33333333333333}{\strut it}   \endgroup} &  \\ 
\hline
BERT-MRP & {\begingroup\setlength{\fboxsep}{1pt} \colorbox{red!2}{\strut all} \colorbox{red!5}{\strut my} \colorbox{red!2}{\strut friends} \colorbox{red!0}{\strut and} \colorbox{red!1}{\strut peers} \colorbox{red!2}{\strut are} \colorbox{red!6}{\strut being} \colorbox{red!11}{\strut openly} \colorbox{red!57}{\strut racist} \colorbox{red!19}{\strut towards} \colorbox{red!57}{\strut asian} \colorbox{red!6}{\strut \#\#s} \colorbox{red!10}{\strut and} \colorbox{red!6}{\strut this} \colorbox{red!100}{\strut bitch} \colorbox{red!13}{\strut is} \endgroup} & OF \\ 
 & {\begingroup\setlength{\fboxsep}{1pt}
\colorbox{red!6}{\strut not} \colorbox{red!11}{\strut having} \colorbox{red!6}{\strut it}   \endgroup} &  \\
BERT-RP & {\begingroup\setlength{\fboxsep}{1pt} \colorbox{red!1}{\strut all} \colorbox{red!2}{\strut my} \colorbox{red!0}{\strut friends} \colorbox{red!2}{\strut and} \colorbox{red!0}{\strut peers} \colorbox{red!1}{\strut are} \colorbox{red!4}{\strut being} \colorbox{red!3}{\strut openly} \colorbox{red!8}{\strut racist} \colorbox{red!8}{\strut towards} \colorbox{red!47}{\strut asian} \colorbox{red!8}{\strut \#\#s} \colorbox{red!12}{\strut and} \colorbox{red!24}{\strut this} \colorbox{red!100}{\strut bitch} \colorbox{red!4}{\strut is} \endgroup} & OF \\ 
 & {\begingroup\setlength{\fboxsep}{1pt}
\colorbox{red!3}{\strut not} \colorbox{red!1}{\strut having} \colorbox{red!1}{\strut it}   \endgroup} & \\
BERT-HX & {\begingroup\setlength{\fboxsep}{1pt}  \colorbox{red!0}{\strut all} \colorbox{red!0}{\strut my} \colorbox{red!0}{\strut friends} \colorbox{red!0}{\strut and} \colorbox{red!0}{\strut peers} \colorbox{red!20}{\strut are} \colorbox{red!0}{\strut being} \colorbox{red!0}{\strut openly} \colorbox{red!0}{\strut racist} \colorbox{red!0}{\strut towards} \colorbox{red!39}{\strut asian} \colorbox{red!39}{\strut \#\#s} \colorbox{red!0}{\strut and} \colorbox{red!23}{\strut this} \colorbox{red!100}{\strut bitch} \colorbox{red!0}{\strut is} \endgroup} & OF \\ 
 & {\begingroup\setlength{\fboxsep}{1pt}
\colorbox{red!19}{\strut not} \colorbox{red!0}{\strut having} \colorbox{red!0}{\strut it}   \endgroup} &  \\
\hline

Human & {\begingroup\setlength{\fboxsep}{1pt} \colorbox{red!0.0}{\strut hey} \colorbox{red!66.66666666666666}{\strut nazi} \colorbox{red!66.66666666666666}{\strut dirtbag} \colorbox{red!66.66666666666666}{\strut is} \colorbox{red!66.66666666666666}{\strut your} \colorbox{red!66.66666666666666}{\strut black} \colorbox{red!66.66666666666666}{\strut bull} \colorbox{red!66.66666666666666}{\strut keeping} \colorbox{red!66.66666666666666}{\strut you} \colorbox{red!66.66666666666666}{\strut occupied}   \endgroup} & OF \\ 
\hline
BERT-MRP & {\begingroup\setlength{\fboxsep}{1pt} \colorbox{red!11}{\strut hey} \colorbox{red!52}{\strut nazi} \colorbox{red!40}{\strut dirt} \colorbox{red!44}{\strut \#\#bag} \colorbox{red!4}{\strut is} \colorbox{red!15}{\strut your} \colorbox{red!31}{\strut black} \colorbox{red!100}{\strut bull} \colorbox{red!0}{\strut keeping} \colorbox{red!0}{\strut you} \colorbox{red!7}{\strut occupied}   \endgroup} & OF \\
BERT-RP & {\begingroup\setlength{\fboxsep}{1pt} \colorbox{red!9}{\strut hey} \colorbox{red!24}{\strut nazi} \colorbox{red!24}{\strut dirt} \colorbox{red!23}{\strut \#\#bag} \colorbox{red!1}{\strut is} \colorbox{red!0}{\strut your} \colorbox{red!25}{\strut black} \colorbox{red!100}{\strut bull} \colorbox{red!16}{\strut keeping} \colorbox{red!0}{\strut you} \colorbox{red!1}{\strut occupied}   \endgroup} & OF \\
BERT-HX & {\begingroup\setlength{\fboxsep}{1pt}  \colorbox{red!36}{\strut hey} \colorbox{red!100}{\strut nazi} \colorbox{red!72}{\strut dirt} \colorbox{red!72}{\strut \#\#bag} \colorbox{red!24}{\strut is} \colorbox{red!0}{\strut your} \colorbox{red!0}{\strut black} \colorbox{red!0}{\strut bull} \colorbox{red!0}{\strut keeping} \colorbox{red!22}{\strut you} \colorbox{red!19}{\strut occupied}   \endgroup} & OF \\
\hline

Human & {\begingroup\setlength{\fboxsep}{1pt} \colorbox{red!66.66666666666666}{\strut white} \colorbox{red!66.66666666666666}{\strut bitch} \colorbox{red!66.66666666666666}{\strut pink} \colorbox{red!66.66666666666666}{\strut pussy}   \endgroup} & OF \\ 
\hline
BERT-MRP & {\begingroup\setlength{\fboxsep}{1pt} \colorbox{red!22}{\strut white} \colorbox{red!100}{\strut bitch} \colorbox{red!0}{\strut pink} \colorbox{red!7}{\strut pussy}   \endgroup} & OF \\
BERT-RP & {\begingroup\setlength{\fboxsep}{1pt} \colorbox{red!0}{\strut white} \colorbox{red!100}{\strut bitch} \colorbox{red!3}{\strut pink} \colorbox{red!2}{\strut pussy}   \endgroup} & OF \\
BERT-HX & {\begingroup\setlength{\fboxsep}{1pt}  \colorbox{red!0}{\strut white} \colorbox{red!100}{\strut bitch} \colorbox{red!0}{\strut pink} \colorbox{red!42}{\strut pussy}   \endgroup} & OF \\

\hline
\end{tabular}
\caption{The highlighted words of the human rationale and the rationale of the models with detection results in 'offensive' sentences. }
\label{offensive_appendix}
}
\end{table*}

\end{document}